\documentclass[11pt,a4paper]{article}
\usepackage[hyperref]{acl2021}
\usepackage{times}
\usepackage{latexsym}
\usepackage{multirow}

\usepackage{microtype}
\usepackage{graphicx}
\usepackage{booktabs}
\usepackage{subcaption}
\usepackage{float}
\usepackage{caption}
\usepackage{subcaption}
\aclfinalcopy 

\title{Language Tags Matter for Zero-Shot Neural Machine Translation}

\author{Liwei Wu, Shanbo Cheng, Mingxuan Wang, Lei Li \\
  ByteDance AI Lab \\
  \textit{\{wuliwei.000,chengshanbo,wangmingxuan.89,lileilab\}@bytedance.com}
}

\date{}

\begin{document}
\maketitle
\begin{abstract}
Multilingual Neural Machine Translation (MNMT) has aroused widespread interest due to its efficiency. An exciting advantage of MNMT models is that they could also translate between unsupervised (zero-shot) language directions. Language tag (LT) strategies are often adopted to indicate the translation directions in MNMT.
In this paper, we demonstrate that the LTs are not only indicators for translation directions but also crucial to zero-shot translation qualities. Unfortunately, previous work tends to ignore the importance of LT strategies. We demonstrate that a proper LT strategy could enhance the consistency of semantic representations and alleviate the off-target issue in zero-shot directions. Experimental results show that by ignoring the source language tag (SLT) and adding the target language tag (TLT) to the encoder, the zero-shot translations could achieve a +8 BLEU score difference over other LT strategies in IWSLT17, Europarl, TED talks translation tasks.
\end{abstract}

\section{Introduction}

\begin{table*}[htb]

\centering
\begin{tabular}{c|c|c}
\toprule
Strategy  &  Source sentence & Target sentence \\
\midrule
Original  &  \texttt{Hello World!} & \texttt{¡Hola Mundo!} \\
\midrule
T-ENC  & \texttt{\textbf{\_\_es\_\_} Hello World!}  & \texttt{¡Hola Mundo!} \\
T-DEC  & \texttt{Hello World!}  & \texttt{\textbf{\_\_es\_\_} ¡Hola Mundo!} \\
S-ENC-T-ENC  & \texttt{\textbf{\_\_en\_\_ \_\_es\_\_} Hello World!} & \texttt{¡Hola Mundo!} \\
S-ENC-T-DEC  & \texttt{\textbf{\_\_en\_\_} Hello World!}  & \texttt{\textbf{\_\_es\_\_} ¡Hola Mundo!} \\
\bottomrule

\end{tabular}
\caption{Examples of modified input data by different LT strategies. The bold tokens are the SLT (\texttt{\_\_en\_\_}) or TLT (\texttt{\_\_es\_\_}). T-ENC is identical to~\cite{DBLP:journals/tacl/JohnsonSLKWCTVW17}, which adds the TLT to the encoder (source) side. T-DEC means placing the TLT on the decoder (target) side of model. S-ENC-T-ENC and S-ENC-T-DEC place the SLT on the encoder side, but the former also places the TLT on encoder side, while the latter on the decoder side.}
\label{table:li_strategies}
\end{table*}
Neural Machine Translation (NMT) based on the \textit{encoder-decoder framework with attention mechanism}~\cite{DBLP:conf/nips/SutskeverVL14,DBLP:journals/corr/BahdanauCB14,DBLP:conf/emnlp/LuongPM15,DBLP:journals/corr/WuSCLNMKCGMKSJL16,gehring2017convolutional,DBLP:conf/nips/VaswaniSPUJGKP17} has achieved state-of-the-art (SotA) results in many language pairs \cite{deng2018alibaba,barrault2019findings}. Pioneered by \cite{dong2015multi,DBLP:conf/naacl/FiratCB16,DBLP:conf/naacl/ZophK16,ha2016toward,DBLP:journals/tacl/JohnsonSLKWCTVW17}, researchers start to investigate the possibility of using a single model to translate between multiple languages, which is the Multilingual Neural Machine Translation (MNMT). Benefiting from the transferring ability of multilingual modeling, MNMT could achieve better translation quality between low-resource language directions than bilingual models~\cite{DBLP:conf/naacl/GuHDL18,DBLP:conf/iclr/WangPAN19}. More exciting, MNMT could even translate between zero-shot language directions~\cite{DBLP:journals/tacl/JohnsonSLKWCTVW17,DBLP:conf/acl/GuWCL19,DBLP:conf/wmt/PhamNHW19,DBLP:conf/emnlp/KuduguntaBCF19}.

Unlike bilingual NMT, language-specific signals should be accessible to the MNMT model so that the model can distinguish the translation directions. Ha et al.,~\shortcite{ha2016toward} first introduced a universal encoder-decoder framework for MNMT models with language-specific coded vocabulary to indicate different languages. The \textit{encoder-decoder} architecture is identical to bilingual models~\cite{DBLP:journals/corr/BahdanauCB14,DBLP:conf/nips/VaswaniSPUJGKP17}. To further simplify the MNMT models, Johnson et al.,~\shortcite{DBLP:journals/tacl/JohnsonSLKWCTVW17} propose to add language tags (LTs) to the beginning of input data to indicate the target language. Then a shared vocabulary could be learned for all languages. The training data of different languages could thus be mixed-up to train the MNMT model. Such a strategy greatly simplifies the training and decoding procedure. We call it the \textit{LT strategy}. 
This paper focuses on investigating the impact of LT strategies for zero-shot translation directions in MNMT (zero-shot MNMT). We conduct translation experiments (Section~\ref{sec:experiments}) and visualization analysis (Section~\ref{sec:visualization}) on several multilingual benchmarks with different LT strategies. We observe that:
\begin{itemize}
    \item The TLT is more important than the SLT. The SLT even causes negative effects on the zero-shot translation.
    \item The placements of LTs have a surprisingly large impact on the translation quality. Placing different LTs on different parts of the NMT model lead to a +8 BLEU score difference in our experiments.
\end{itemize}

Our contributions are mainly twofold: (i) We find that the LT strategies are crucial for the zero-shot MNMT translation quality. Ignoring SLTs and placing the TLTs on the encoder side could achieve the best performance during our experiments. (ii) We conduct extensive visualization analysis to demonstrate that the proper LT strategy could enhance the consistency of semantic representation and alleviate the off-target issue~\cite{DBLP:conf/acl/ZhangWTS20}, thus improving the translation quality. To the best of our knowledge, this is the first paper to systematically study the importance of LT strategies for zero-shot translation quality.

\section{Background and Notations}

Improving the consistency of semantic representations and alleviating the off-target issue~\cite{DBLP:conf/acl/ZhangWTS20} are effective ways to improve the zero-shot translation quality~\cite{al2019consistency,arivazhagan2019missing,zhu2020language}. The semantic representations of different languages should be close to each other to get better translation quality~\cite{ding2017visualizing}. The off-target issue indicates that the MNMT model tends to translate input sentences to the wrong languages, which leads to low translation quality.

Due to its simplicity and efficiency, LT strategy has become a fundamental strategy for MNMT~\cite{dabre2020a}. Though previous work adopted different LT strategies~\cite{wang2018three,DBLP:conf/coling/BlackwoodBW18,DBLP:conf/nips/ConneauL19,liu2020multilingual}, the usages of LT strategies are intuitive and lack systematic study. In this paper, we investigate 4 popular LT strategies, namely \texttt{T-ENC}, \texttt{T-DEC}, \texttt{S-ENC-T-ENC} and \texttt{S-ENC-T-DEC}. Each of them only requires simple modifications to the input data. Table~\ref{table:li_strategies} comprehensively illustrates the strategies with an English to Spanish translation pair (\texttt{Hello World!} $\rightarrow$ \texttt{¡Hola Mundo!}).

\section{Experiments}

\label{sec:experiments}
\begin{table*}[htb!]

\centering
\begin{tabular}{c|c|c|c|c}
\toprule
\multirow{2}{*}{} &                 &  \textbf{\#zero-shot}         & \textbf{\#training sents}  & \textbf{\#sents}\\
\textbf{Dataset} & \textbf{languages} & \textbf{directions} & \textbf{per direction}  & \textbf{per testset}\\ 
\midrule
\textbf{IWSLT17} & en, it, nl, ro & 6 & 145k & 1144 \\
\midrule
\textbf{Europarl} & en, fr, de, es & 6 & 1.96m  & 2000 \\
\midrule
\multirow{3}{*}{\textbf{TED}}  & en, ar, he, ru, ko, it, ja & &  & \\
                      & zh, es, fr, pt, nl, tr, ro & 342 & 187k & 4507 \\
                      & pl, bg, vi, de, fa, hu & &  & \\
\bottomrule

\end{tabular}
\caption{An overview of the datasets. The second column is the languages the training data contains. The third column denotes the number of zero-shot translation directions. The fourth and fifth column denote the averaged number of training data and test data per language direction, respectively.}
\label{table:dataset_description}
\end{table*}

\begin{table*}[htb!]
\centering
\begin{tabular}{@{}lllll@{}}
\toprule
\textbf{Dataset}                    & \textbf{LT Strategy} & \textbf{Supervised} & \textbf{Zero-Shot}      & \textbf{Off-Target (\%)} \\ \midrule
\multirow{4}{*}{\textbf{IWSLT17}}   & T-ENC                 & 32.30               & \textbf{16.00} (+14.02) & \textbf{9.16}            \\
                                    & T-DEC                 & 32.43               & 10.44                   & 29.50                    \\
                                    & S-ENC-T-ENC           & 32.56               & 1.98                    & 94.14                    \\
                                    & S-ENC-T-DEC           & 32.39               & 7.67                    & 48.87                    \\ \midrule
\multirow{4}{*}{\textbf{Europarl}}  & T-ENC                 & 35.55               & \textbf{32.25} (+24.24) & 1.18                     \\
                                    & T-DEC                 & 35.49               & 30.73                   & \textbf{1.13}            \\
                                    & S-ENC-T-ENC           & 35.53               & 8.01                    & 79.53                    \\
                                    & S-ENC-T-DEC           & 35.53               & 29.81                   & 2.26                     \\ \midrule
\multirow{4}{*}{\textbf{TED talks}} & T-ENC                 & 25.63               & \textbf{10.69} (+8.78)  & \textbf{12.63}           \\
                                    & T-DEC                 & 25.58               & 3.11                    & 58.47                    \\
                                    & S-ENC-T-ENC           & 25.84               & 4.07                    & 65.03                    \\
                                    & S-ENC-T-DEC           & 25.63               & 1.91                    & 77.02                    \\ \bottomrule
\end{tabular}
\caption{Translation results on 3 datasets. The \textbf{supervised} and \textbf{zero-shot} column denote the averaged BLEU score of supervised or zero-shot directions. The \textbf{off-target (\%)} denotes the averaged percentage of sentences being translated to wrong languages in zero-shot directions.}
\label{table:translation_results}
\end{table*} 
\subsection{Experiment Settings}
\label{subsec:settings}
\textbf{Datasets} We carry out our experiments on the publicly available IWSLT17~\cite{cettolo2017overview}, TED talks~\cite{DBLP:conf/naacl/QiSFPN18} and Europarl v7~\cite{koehn2005europarl} datasets. Table~\ref{table:dataset_description} shows an overview of the datasets. We choose four different languages (English included) for both IWSLT17 and Europarl, and 20 languages for TED talks. All the training data are English-centric parallel data, which means either the source-side or target-side of the sentence pair is English. We have 6, 6, and 342 zero-shot translation directions and an average of 145k, 1.96M (M = million), and 187k sentence pairs per direction for the three datasets respectively. We choose the official tst2017, WMT \texttt{newstest08}, and the TED talks testsets~\cite{DBLP:conf/naacl/QiSFPN18} as our test sets, respectively.
We learned a joint SentencePiece model~\cite{kudo2018sentencepiece} for sub-word training on all languages with 40,000 merge operations for each dataset. We limit the size of joint vocabulary to 40,000 for all three datasets. \newline
\textbf{Settings} We use the open-source implementation~\cite{DBLP:conf/naacl/OttEBFGNGA19} of Transformer model~\cite{DBLP:conf/nips/VaswaniSPUJGKP17}. Following the settings of~\cite{DBLP:journals/corr/abs-2012-15127}, we use a $5$-layer encoder and $5$-layer decoder variation of Transformer-base model~\cite{DBLP:conf/nips/VaswaniSPUJGKP17} for TED and IWSLT17. For Europarl v7, we use a standard Transformer-big model~\cite{DBLP:conf/nips/VaswaniSPUJGKP17}. Sentence pairs are batched together by approximate sentence length. Each batch has approximately 30,000 source tokens and 30,000 target tokens.
We use the Adam~\cite{Kingma2015Adam} optimizer to update the parameters and train each model for 100,000 steps to make sure it converges. We use beam search for heuristic decoding, and set the beam size to 4. We use SacreBLEU~\cite{papineni2002bleu,post2018call} to evaluate the translation results. To calculating the percentage of off-target translations, we use the langdetect\footnote{https://github.com/Mimino666/langdetect} tool to detect the language of the translated sentences.
\subsection{Experimental Results}
\label{subsec:results}
We show the translation results on the IWSLT17, Europarl, and TED talks datasets in Table~\ref{table:translation_results}. For all three datasets, different strategies achieve comparable BLEU score on supervised directions. However, for the zero-shot directions, the BLEU score varies significantly using different LT strategies. 
One observation is that the \texttt{T-ENC} strategy consistently outperforms the other three strategies on all datasets in terms of BLEU score with large margin, regardless of the corpus size and number of languages. In terms of off-target issue, \texttt{T-ENC} achieves the best performance in most cases. 

Besides, ignoring the SLT (\texttt{T-ENC} v.s.~\texttt{S-ENC-T-ENC}) also helps the zero-shot BLEU score. The percentage of off-target translations reaches 94.14\% in the \textbf{IWSLT17} dataset by \texttt{S-ENC-T-ENC} strategy, while only 9.16\% by \texttt{T-ENC} strategy. It indicates that the model translates almost all the sentences to the wrong languages in \texttt{S-ENC-T-ENC}, while to the right languages in \texttt{T-ENC}. It proves again the SLT hurts the zero-shot translation.

Another interesting observation is that placing the TLT on the encoder side also helps the zero-shot performance. Compared with \texttt{T-ENC}, both the translation quality and off-target performance are significantly worse in \texttt{T-DEC}. We will study the reasons behind the above observations by visualization analysis in Section~\ref{sec:visualization}.

\section{Visualization Analysis}

 \label{sec:visualization}
We conduct the visualizations on the TED talks data to analyze the impact of different LT strategies on the semantic representation consistency and the off-target issue in MNMT.
\begin{figure}[htb!]
\begin{subfigure}{.23\textwidth}
  \centering
  \includegraphics[width=\linewidth]{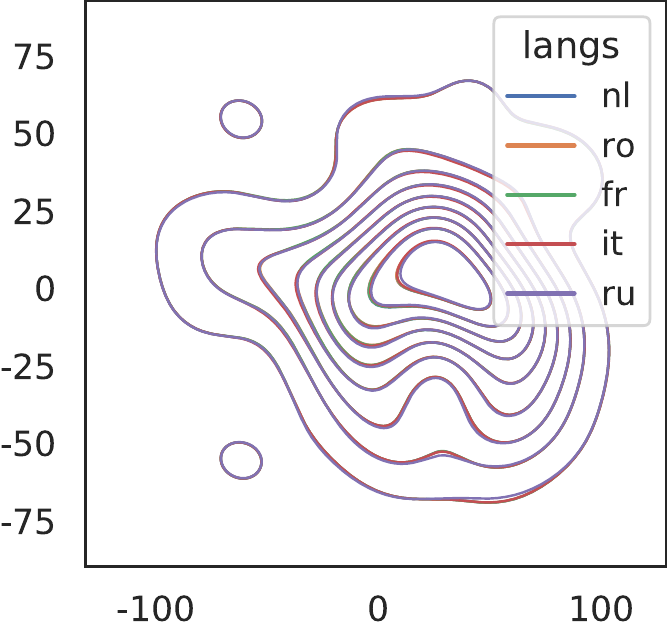}  
  \caption{T-ENC}
  \label{fig:kde-t-enc}
\end{subfigure}
\begin{subfigure}{.23\textwidth}
  \centering
  \includegraphics[width=\linewidth]{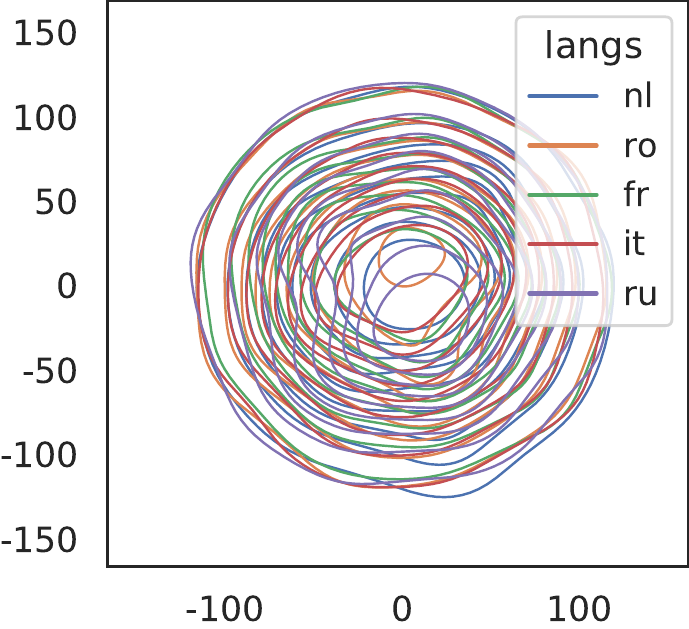}  
  \caption{T-DEC}
  \label{fig:kde-t-dec}
\end{subfigure}

\begin{subfigure}{.23\textwidth}
  \centering
  \includegraphics[width=\linewidth]{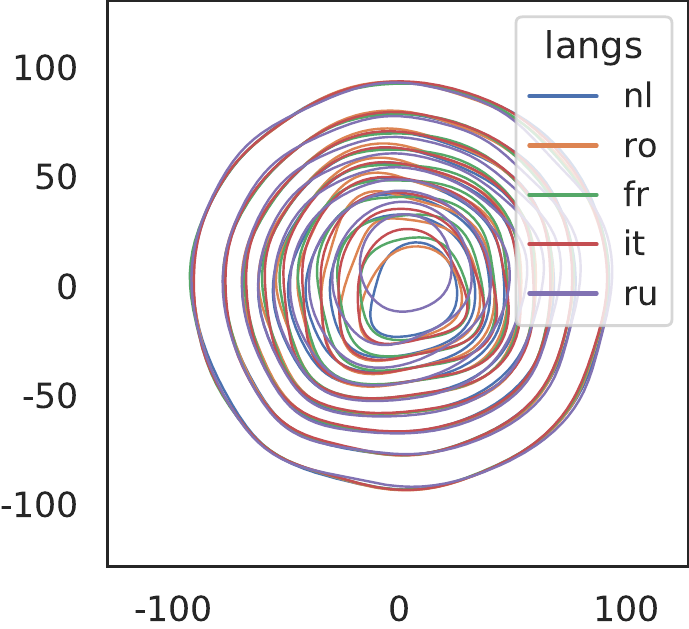}  
  \caption{S-ENC-T-ENC}
  \label{fig:kde-s-enc-t-enc}
\end{subfigure}
\begin{subfigure}{.23\textwidth}
  \centering
  \includegraphics[width=\linewidth]{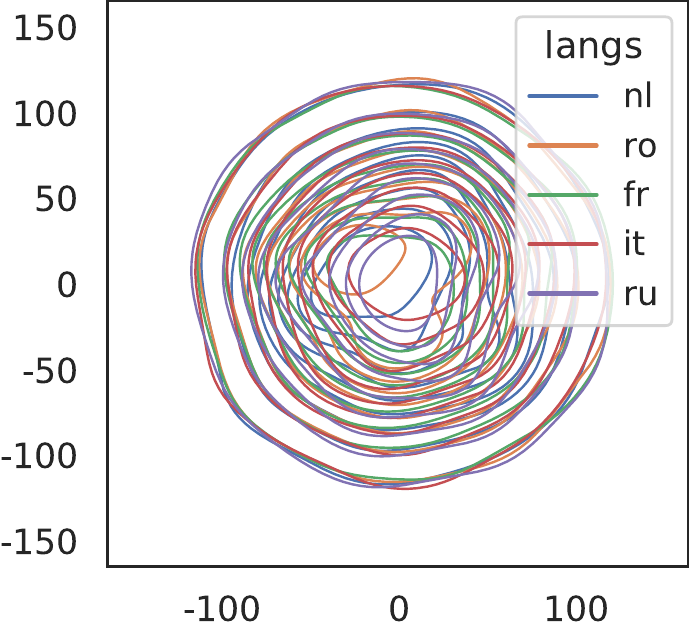}  
  \caption{S-ENC-T-DEC}
  \label{fig:kde-s-enc-t-dec}
\end{subfigure}
\caption{KDE visualization of encoder output using different LT strategies. 5 languages (nl, ro, fr, it, ru $\,\to\,$ zh) are randomly chosen for better readability. }
\label{fig:kde_plot}
\end{figure}
\paragraph{Enhancing the Semantic Representation Consistency} Figure~\ref{fig:kde_plot} shows the kernel density estimation (KDE)~\cite{parzen1962estimation} of t-SNE~\cite{van2008visualizing} reduced average encoder output on different languages. We randomly chose 5 source languages (nl, ro, fr, it, ru $\,\to\,$ zh) instead of all languages for clearer visualization. We choose 100 sentences for each language, and each sentence has its corresponding translation in the other 4 languages. The contour lines drawn by Kernel Density Estimation tools \footnote{https://seaborn.pydata.org/generated/seaborn.kdeplot.html} was used to estimate the semantic distribution of the encoder outputs. The contour lines visualize the semantic representation of different languages. The representations are more consistent if the contour lines of different languages overlap more with each other.

The contour lines are nearly perfectly overlap with each other in \texttt{T-ENC} (Figure~\ref{fig:kde-t-enc}), while they do not for the other strategies. Comparing Figure~\ref{fig:kde-t-enc} and Figure~\ref{fig:kde-s-enc-t-enc}, we can see that ignoring SLT greatly helps the model to learn more consistent representations. Comparing Figure~\ref{fig:kde-t-enc} and Figure~\ref{fig:kde-t-dec}, placing TLT to encoder side instead of the decoder side also helps the semantic consistency. Both comparisons validate that \texttt{T-ENC} could learn the most consistent and different semantic representations, thus achieves the best BLEU score. It might be why the shape of contour lines in \texttt{T-ENC} is significantly different from other strategies. 

\begin{figure}[htb!]
\begin{subfigure}{.5\textwidth}
    \begin{subfigure}{.47\textwidth}
      \centering
      \renewcommand\thesubfigure{\alph{subfigure}1}
      \includegraphics[width=\linewidth]{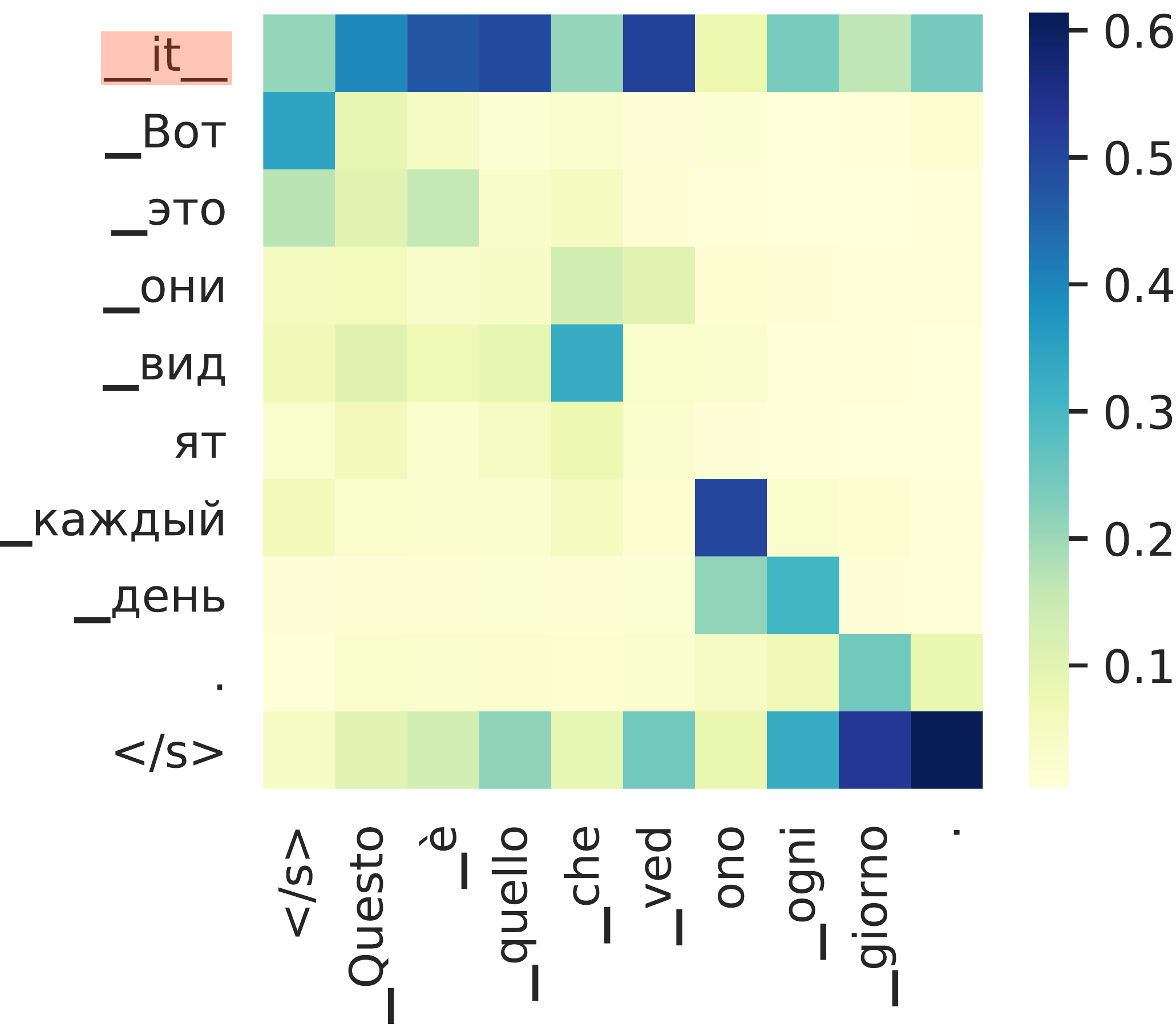}  
      \caption{T-ENC}
      \label{fig:t_enc_att}
    \end{subfigure}
    \begin{subfigure}{.47\textwidth}
      \centering
      \addtocounter{subfigure}{-1}
      \renewcommand\thesubfigure{\alph{subfigure}2}
      \includegraphics[width=\linewidth]{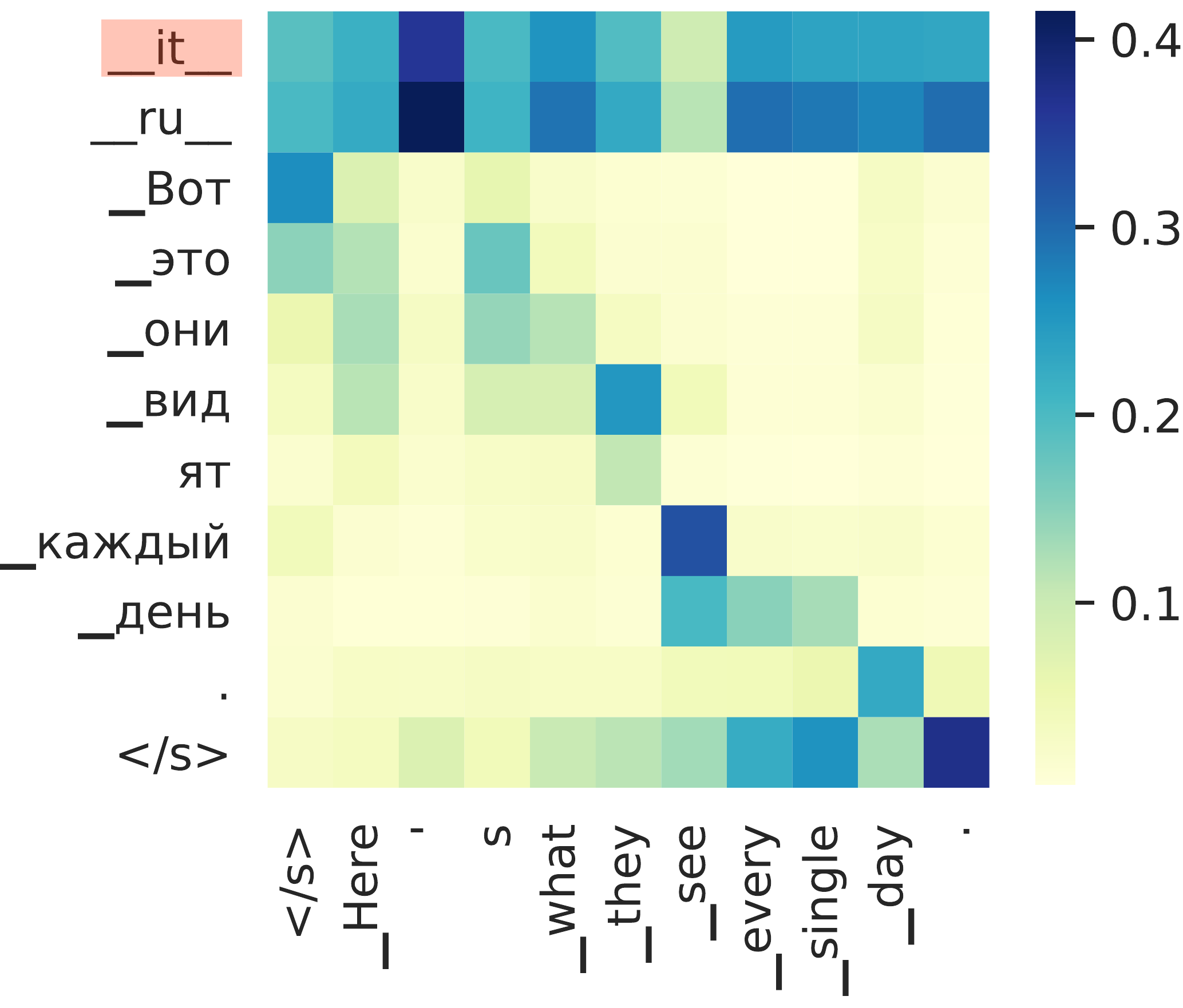}  
      \caption{S-ENC-T-ENC}
      \label{fig:s_enc_t_enc_att}
    \end{subfigure}
\addtocounter{subfigure}{-1}
\caption{The Decoder Cross Attention}
\end{subfigure}

\begin{subfigure}{.5\textwidth}
    \begin{subfigure}{.47\textwidth}
      \centering
      \renewcommand\thesubfigure{\alph{subfigure}1}
      \includegraphics[width=\linewidth]{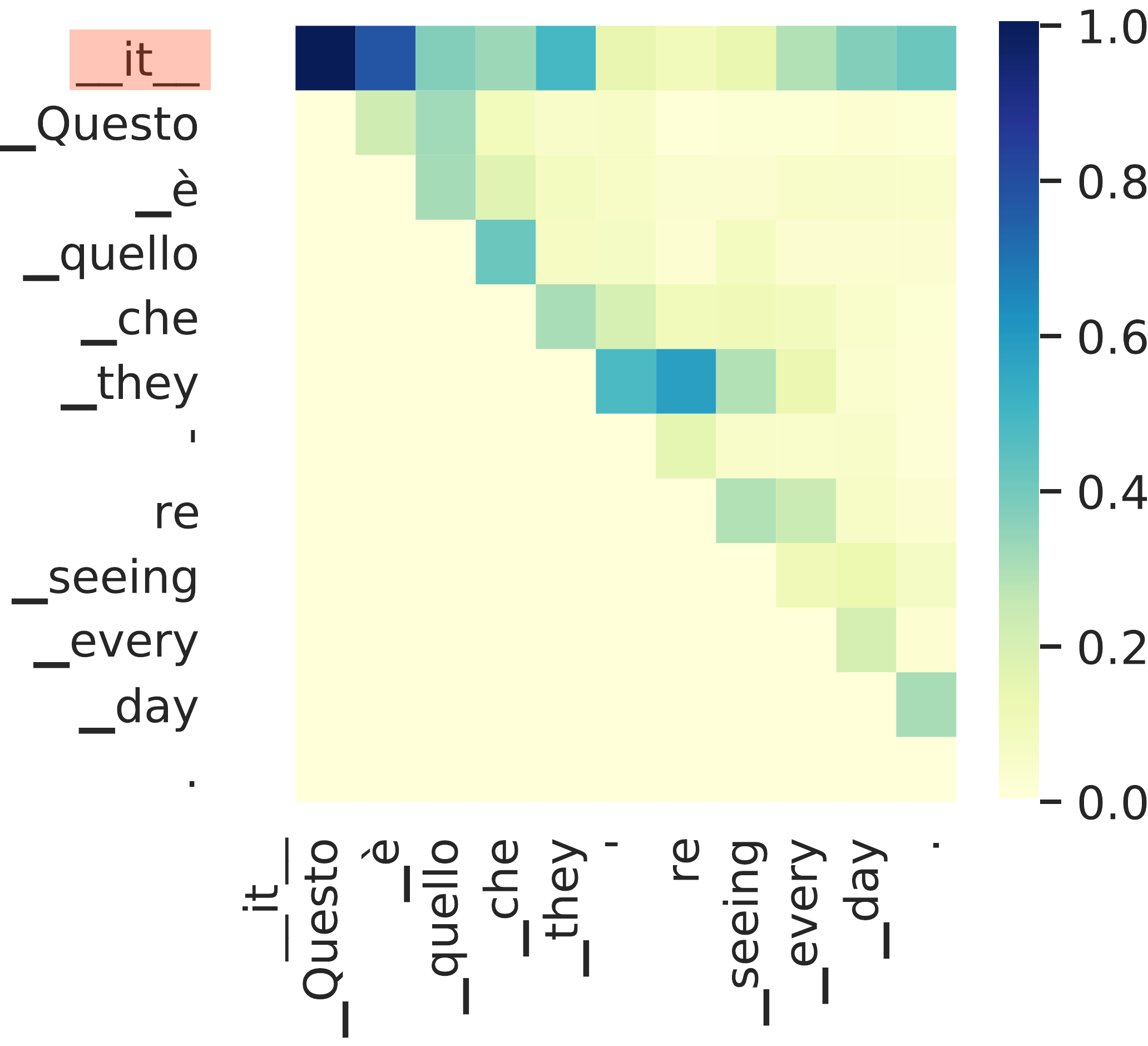}  
      \caption{T-DEC}
      \label{fig:t_dec_att}
    \end{subfigure}
    \begin{subfigure}{.47\textwidth}
      \centering
      \addtocounter{subfigure}{-1}
      \renewcommand\thesubfigure{\alph{subfigure}2}
      \includegraphics[width=\linewidth]{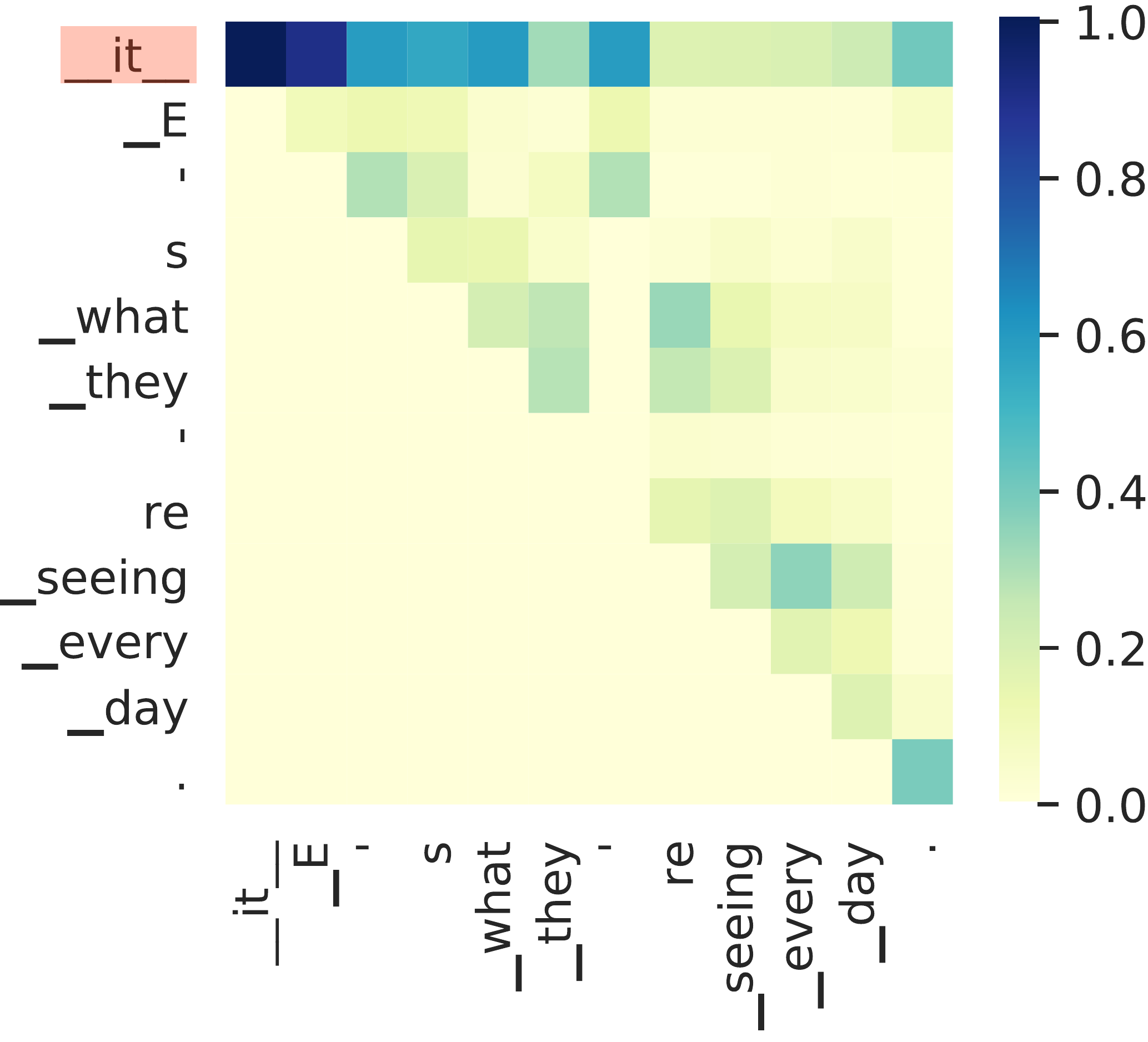}  
      \caption{S-ENC-T-DEC}
      \label{fig:s_enc_t_dec_att}
    \end{subfigure}
\addtocounter{subfigure}{-1}
\caption{The Decoder Self Attention}
\end{subfigure}
\caption{Attention visualization on a Russian to Italian translation example using different LT strategies. Note that we present the cross-attention for \texttt{T-ENC} and \texttt{S-ENC-T-ENC}, the decoder self-attention for \texttt{T-DEC} and \texttt{S-ENC-T-DEC} to visualize the TLT token.}
\label{fig:attention}
\end{figure}

\paragraph{Alleviating the off-target Issue}
Figure~\ref{fig:attention} shows the attention visualization of a Russian to Italian translation example using different LT strategies. The x-axis is the Italian translation.

In Figure~\ref{fig:t_enc_att}, \texttt{T-ENC} strategy pays attention to the TLT (in this case, the token \texttt{\_\_it\_\_} in the red background) during the whole translation procedure (left-to-right). Compared to \texttt{T-ENC}, both \texttt{T-DEC} and \texttt{S-ENC-T-DEC} pay less attention to the TLT after a few tokens are generated. It validates that placing the TLT on the encoder side would also help the model distinguish the target languages. The \texttt{S-ENC-T-ENC} pays nearly equal attention to both SLT and TLT, which might make the model confused about which one is the target language. Both comparisons prove that the \texttt{T-ENC} strategy has the best ability to distinguish the target languages, thus alleviates the off-target issue.

\begin{figure}[htb!]
\centering
\begin{subfigure}{0.23\textwidth}
  \centering
  \includegraphics[width=\linewidth]{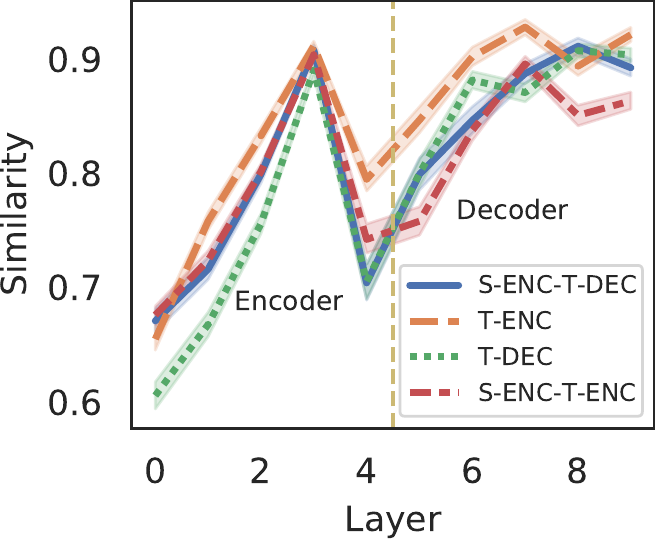}  
  \caption{many-to-one}
  \label{fig:many_to_one}
\end{subfigure}
\begin{subfigure}{.23\textwidth}
  \centering
  \includegraphics[width=\linewidth]{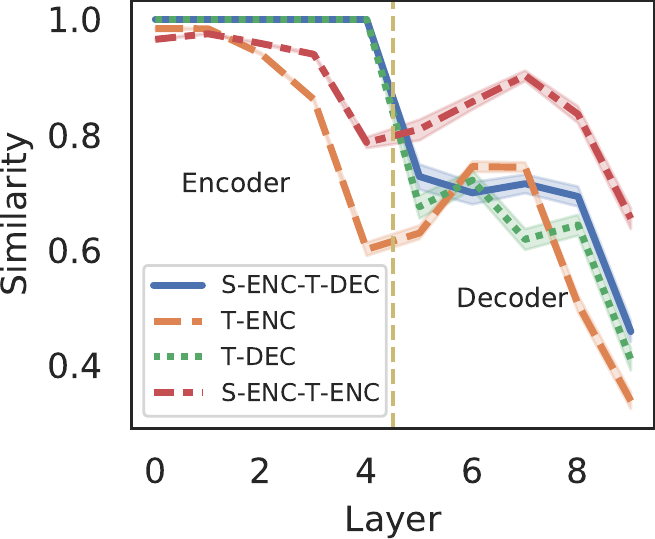}  
  \caption{one-to-many}
  \label{fig:one_to_many}
\end{subfigure}
\caption{The cosine similarity of layer-wise output using different LT strategies. Note that layer 0 to 4 are the encoder layers, layer 5 to 9 are the decoder layers.}
\label{fig:similarity}
\end{figure}

\paragraph{Combining Both Semantic Consistency and off-target Issue} 
Figure~\ref{fig:similarity} visualizes the cosine similarity of the layer-wise encoder and decoder output of different languages in zero-shot setting (English excluded). We sampled 100 muti-way data from the test set and averaged the cosine similarity between each language.
 
In the many-to-one setting, we randomly select Russian as the target language and translate the other 18 languages to Russian to obtain the model outputs. The similarity improves from encoder layer 0 to 3 and decoder layer 0 to layer 4, which indicates that the semantic consistency improves as the layer goes up. Interestingly, the similarity drops from encoder layer 3 to layer 4. It might be because the decoder interacts with the encoder directly between encoder layer 4 and decoder layer 0, thus interferes with the top-layer encoder output. But the dropping trend is less rapid in \texttt{T-ENC} than in other strategies. 
The \texttt{T-ENC} achieves the highest similarity on the last layer, which shows that the \texttt{T-ENC} learns more consistent semantics representations. 

In the one-to-many setting, we treat Russian as the source language and translate Russian to the other 18 languages to get the model output. The semantic similarity drops as the layer goes up in all four strategies. It indicates that the model can distinguish different target languages as the layer goes up. \texttt{T-ENC} achieves the lowest similarity at the last layer output among all strategies. It shows again that the \texttt{T-ENC} has the best ability to alleviate the off-target issue.

\section{Conclusion}

We show that the language tags in MNMT are not just indicators for translation directions but also significantly impact the zero-shot translation quality. By extensive experiments and visualization analysis, we found that (i) ignoring the SLTs could help the models learn consistent semantic representations. (ii) Placing the TLTs on the encoder side could help the decoder pay more attention to the target language, thus alleviating the off-target issue. Zero-shot translation quality could be improved by investigating how to enhance the semantic representation consistency further and alleviate the off-target issue by optimizing LT strategies. We will conduct methods to optimize the LT strategy in our future work.

\bibliographystyle{acl_natbib}
\bibliography{anthology,new}


\end{document}